\documentclass[default,iicol]{sn-jnl}
\normalbaroutside

\jyear{2023}%

\theoremstyle{thmstyleone}%
\theoremstyle{thmstyletwo}%

\theoremstyle{thmstylethree}%

\usepackage{todonotes}
\usepackage{siunitx}
\usepackage{subcaption}
\usepackage{graphicx}

\begin{document}

\title[Article Title]{Steel Surface Roughness Parameter Calculations Using Lasers and Machine Learning Models}


\author*[1]{\fnm{Alex} \sur{Milne}}\email{alexander.milne@hotmail.co.uk}

\author[1]{\fnm{Xianghua} \sur{Xie}}\email{X.Xie@swansea.ac.uk}

\affil[1]{\orgdiv{Department of Computer Science}, \orgname{Swansea University}, \orgaddress{\city{Swansea}, \country{UK}}}


\abstract{

Control of surface texture in strip steel is essential to meet customer requirements during galvanizing and temper rolling processes. Traditional methods rely on post-production stylus measurements, while on-line techniques offer non-contact and real-time measurements of the entire strip. However, ensuring accurate measurement is imperative for their effective utilization in the manufacturing pipeline. Moreover, accurate on-line measurements enable real-time adjustments of manufacturing processing parameters during production, ensuring consistent quality and the possibility of closed-loop control of the temper mill.
In this study, we leverage state-of-the-art machine learning models to enhance the transformation of on-line measurements into a significantly more accurate Ra surface roughness metric. By comparing a selection of data-driven approaches, including both deep learning and non-deep learning methods, to the close-form transformation, we evaluate their potential for improving surface texture control in temper strip steel manufacturing.
}

\keywords{Machine Learning, On-line measurement, Surface roughness, Temper rolling, Time Series Extrinsic Regression (TSER)}



\maketitle

\section{Introduction}\label{sec:intro}

Temper rolling is a critical process in strip steel production, which involves cold-rolling the steel to improve its mechanical and surface properties. For high-value products like automotive paneling, meeting customer demands extends beyond achieving favorable mechanical properties and necessitates specific surface properties to facilitate better press performance \cite{RaOnFrictionPress} and ensure paint quality \cite{bastawros1993paineffects}.

Line operators have the ability to modify various parameters, including roll selection, roll force, and speed, in order to influence the steel's microstructure and meet customer requirements. The selection of rolls in the temper mill plays a crucial role in imparting surface texture onto the steel. The rolls have a surface texture created by electric discharged texturing, which is then transferred onto the steel by the roll with respect to the other parameters that affect how this texture transfer happens, such as roll force and speed. 

The feedback process informing line operators about whether the surface texture has met customer specifications is slow.
Surface measurements are taken post-production manually using a stylus device, and only on a small section at the head or tail of the coil. This causes two main issues: a) the samples may not be representative of the entire coil surface~\cite{cheri2018online} b) the feedback does not allow mid-coil process adjustments. This can result in producing coils that do not meet customer requirements.
Therefore, the slow feedback provided is only beneficial for subsequent coils, leaving the inadequate coil with a reduced value and forcing re-production of the product, potentially multiple times in a slow iterative process until the customer requirements are met. Given the prevalence of just-in-time manufacturing \cite{SayerJustInTime}, the delayed feedback-driven re-production can cause downstream delays and the possibility of costly line stoppages for customers if replacement steel is not manufactured and delivered promptly.

Access to fast monitoring on-line allows operators to perform real-time adjustments for line parameters and provides measurements for the entire surface. It also opens the door for real-time closed-loop high-precision control systems to automate the temper mill parameters to create steel for customer requirements. 
Cheri et al.~\cite{OnlineRoughnessControlMethod} propose an on-line intelligent control method of cold rolled strip steel surface roughness, which uses on-line measurements to inform a model outputting roll force for closed-loop roughness control. The authors use a fuzzy neural network for their prediction. This method relies on the accuracy of the on-line measurement system used as this is the value targeted by the system. 

To address the need for on-line measurement, fast non-contact measurement techniques have been developed \cite{OnlineRoughnessControlMethod}. The technique we use fires a laser at the surface and the scattered light intensities are captured giving a corresponding angle. Surface gradients can be calculated from the reflected angles through time and integrated to calculate the surface profile and surface statistics. 

Other techniques exist; for instance, the work by Bilstein et al.~\cite{Two_systems_for_on-line_measurement} provides details of a system that can measure surface roughness on-line and offer information on the measurement principle, system design, and laboratory and on-line trial results of the system. An early attempt is provided by Luk et al.~\cite{measureing_roughness_1989}. The authors use a microscope to capture scattered light which is represented using a gray-level histogram of light intensities from which the optical roughness parameter is calculated. The system is calibrated using parameters calculated using traditional techniques on the sample.

For adjustments to be made based on optical measurements, it is essential that the measurements are accurate. However, when exploring this technique side by side with the traditional stylus-based method, we have found the accuracy of surface statistics calculated not suitable for product release.

We propose the use of machine learning to improve the accuracy and effectiveness of this method by performing the transformation from raw laser reflection data to surface parameters. By accounting for the various unknown factors that influence the physical processes and the raw reflections which impact the resulting surface roughness, machine learning can develop a more accurate model for prediction, which can provide surface roughness in real-time during production. Ultimately, this will improve the quality and efficiency of the temper rolling process. 

In the domain of Brain-Computer Interfaces, several studies~\cite{kollHod2023deep, kang2014bayesian, schirrmeister2017deep} delve into machine learning for electroencephalogram (EEG) datasets primarily aimed at classifying actions based on brain signals, such as distinguishing between left and right hand movements. Although these are classification problems, they share a resemblance with our regression problem, as they involve multichannel signals collected from multiple sensors spanning a substantial number of time steps, contingent upon the measurement duration and frequency.

Our research builds upon existing literature~\cite{OnlineRoughnessControlMethod} by introducing methods to enhance accuracy and transform the Ra parameter to align with the gold standard stylus measurement, as opposed to roughness from another form of surface measurement. This transformation ensures alignment with customer requirements, where the Ra ground truth is defined using a stylus measurement. Ra is defined as the mean deviation of the roughness profile, formalized as follows:
\begin{equation}
R_a = \frac{1}{N} \sum_{i=1}^N | z_i - \overline{z} |
\label{eq:Ra}
\end{equation}

\noindent where $N$ is the number of time steps, $z$ is the height profile, and $\overline{z}$ is the mean of $z$. 

Previous studies have also utilized machine learning to regress the Ra parameter in a production setting. For instance, Elangovan et al. \cite{ELANGOVANRoughnessPredictionInTurning} proposed the use of machine learning to characterize metal surface roughness Ra in the context of metal objects turned on a lathe. Their approach involved employing multiple regression analysis on statistical features extracted from the vibration signal of the tool, alongside machining parameters such as tool wear and speed. The statistical features extracted were relatively straightforward, including measures such as mean and skewness. Notably, the authors did not incorporate any light sensing data of the surface into their model, while in this paper, we exclusively model with light sensing data.

The surface roughness Ra is an important parameter in other fields of manufacturing such as 3D printing~\cite{MUSHTAQ2023108129, MUSHTAQ2023546, MushtaqWangRehmanKhanBaoSharmaEldinAbbas+2023, ma16093392}, where the Ra value can be correlated with other desirable properties. 
In another use case of machine learning for surface roughness prediction~\cite{LAFEPERDOMO2023148}, the authors use a Multi-layer Perceptron (MLP) and adaptive neuro-fuzzy inference system for roughness prediction in stainless steel selective laser melting additive manufacturing.
The Ra parameter has also been used in multiple regression analysis to understand the influence of different parameters in the dressing (replenishing) grinding wheels~\cite{MOHITE2023}. Similarly, Neural Networks (NNs) have been used to model the relationship in Computer Numerical Control (CNC) steel milling between the resulting Ra and material removal rate when using different input parameters~\cite{VISHNUVARDHAN201827058}.

The paper by Gupta et al.~\cite{GUPTA2023112321} presents a machine learning approach for multiclass classification of steel microstructure types, which bears relevance to our work using machine learning. Their research employed the k-nearest neighbor (k-NN) algorithm to train a classifier based on steel composition and heat treatment parameters, analogous to our utilization of machine learning for predictive modeling in the steel manufacturing domain.

In this paper, we introduce an approach that leverages the power of machine learning to substantially improve the on-line measurement of Ra roughness parameters compared to the baseline approach. To our knowledge machine learning has not previously been used for this. Our method transforms raw reflection data into accurate Ra values using machine learning, while the closed-form approach relies on transforming the signal into a surface profile from integrated gradients, from which Ra is calculated. Our research introduces and integrates a diverse set of machine learning architectures from the existing literature and, where required, adapts and optimizes them to suit the intricacies of this problem. By doing so, we not only demonstrate the potential of machine learning in solving this complex transformation task but also provide a methodology and evaluation of these models which can be used by others for similar problems.

\noindent
Contributions of our work:
\begin{itemize}
\item Formulation of the Ra prediction problem as a machine learning problem, including appropriate preparation of the dataset. 
\item Apply machine learning to improve the accuracy of the transformation from laser measurements into the surface roughness Ra parameter. 
\item Give our experimental methodology and provide comprehensive insights into our model training protocols.
\item Provide a comprehensive overview of the various models that can be used to solve time series problems where the channels are spatially related. These are the models we employ in our experiments, categorized into distinct methodological approaches: Non-Deep Learning, 1D Deep Learning, and 2D Deep Learning.
\end{itemize}

The paper is organized as follows. 
Section~\ref{sec:data_stuff} gives an in-depth overview of the data, including: details of the measurement device employed by our industrial collaborator, from which our data is gathered; how we organize the data into a dataset for the machine learning for our experiments; the data measurement methodology; and the issues with measuring ground truth. 
Section \ref{sec:methodology} presents the experimental setup, methodology, model training details and hyperparameters. The section also including a description of the calculation of baseline closed-form solution, and a comprehensive overview of the various data-driven models employed in our experiments categorized into distinct methodological approaches: Non-Deep Learning, 1D Deep Learning, and 2D Deep Learning.
Section \ref{sec:results} presents the results of the data-driven approach experiments vs those of the baseline, and provides additional insights into the generalization ability with given new unseen steel through the results on on the k-fold modelling experiments.
Finally, Section \ref{sec:conclusion} summarizes the key findings of the study and provides recommendations for future work.

\section{Data Acquisition}\label{sec:data_stuff}

\subsection{On-line Measurement Device}\label{sec:online_measuremnt_device}

\begin{figure}[ht]
\centering
\includegraphics[width=\linewidth]{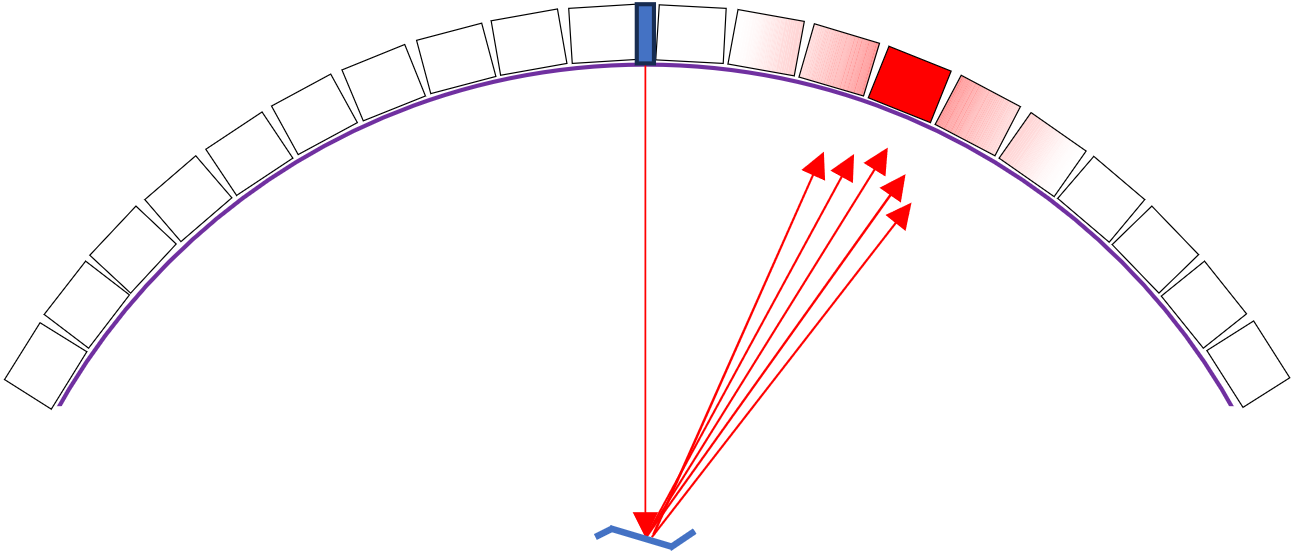}
\caption{Illustrative diagram of the laser measurement device. The intensities of the 20 sensors correspond to a single time step creating a 20x1 1D image in the data plot \ref{fig:zoomed_in_raw_sorm}.}
\label{fig:diagram_of_laser_device}
\end{figure}

The on-line measurement device consists of a laser and a 1D semicircular array of 20 sensors. The laser is positioned between the central two sensors in the array.
The sensors are spaced in 6.7-degree increments, except the central position containing the laser, which is spaced 6.8 degrees, in total spanning an arc of 127.4 degrees. The raw intensity data are generated by firing a continuous laser at the surface of the steel and capturing the reflected light intensities across the array of sensors at discrete intervals. Each discrete measurement creates a 1D image of the angle of reflection and light scattering, with each of the sensors providing a pixel. The measurement device scans the surface of the steel while remaining in a fixed position, as the steel moves through the production line. This means that the measurement is along the rolling direction of the steel strip. The laser pulse rate is synced with the speed of the production line such that each reading is taken at a specified equal distance. This results in intensity-angle information through time, $\mathbb{N}^{C \times T}$, where $C = 20$, the number of sensors; $T$ is the number of time steps; and the values are constrained between 0 and 255. The device calculates the Ra using this data as specified in the baseline calculation section. An illustrative diagram of the device is shown in Fig.~\ref{fig:diagram_of_laser_device} and an example of the intensity data produced by this system can be seen in Fig.~\ref{fig:zoomed_in_raw_sorm}.

\subsection{The Steel Surface dataset}\label{sec:data}

\begin{figure}[ht]
\centering
\includegraphics[width=\linewidth]{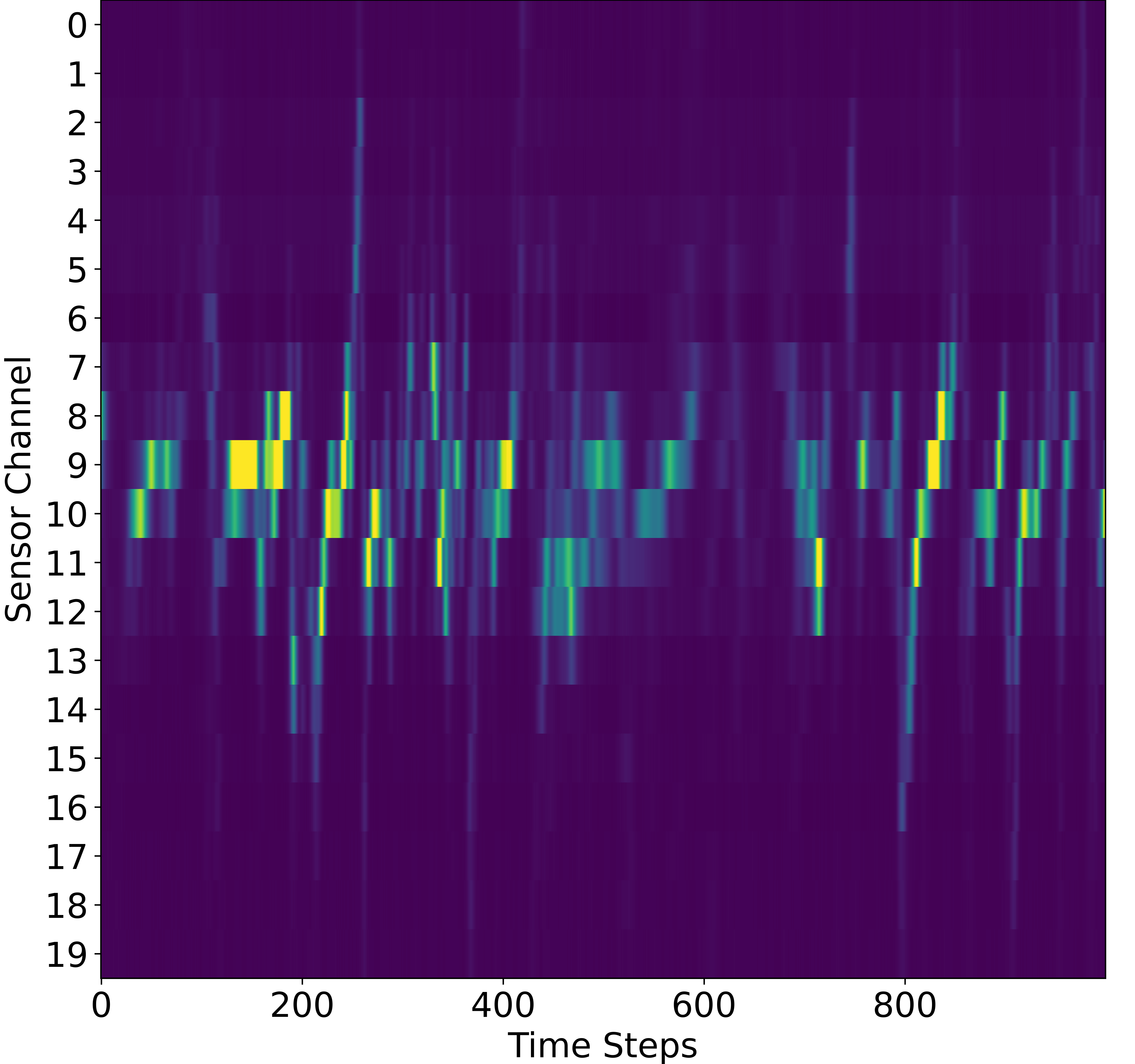}
\caption{The light intensity through time/distance, recorded by the semicircular 20 sensor array collecting light from laser reflected from the steel surface. Each time step is a 20x1 1D image of the angle of reflection, taken with an interval of 0.8um along the length of the steel.}

\label{fig:zoomed_in_raw_sorm}
\end{figure}


The data used in this paper has been provided by our industrial partner for the improvement in Ra calculation from the laser measurement system and to judge the feasibility of a data-driven approach in terms of both speed and accuracy. The data has two measurement types, the conventional stylus measurement, and the laser intensity reflection data. All measurements are taken from a set of 49 steel samples which have been processed differently and have different roughnesses. The samples are size 110x90mm. From each steel sample, multiple measurements are taken in different positions using both measurement types. 3 out of the 49 steel samples use a different coating than the remaining 46 galvanized samples. These samples are expected to be more difficult because they have different reflective properties than the rest of the dataset and fewer samples for the model to learn from.

The laser intensity data for our dataset has a time step distance of 0.8um with each of the readings having $T = 65536$ ($2^{16}$) time steps, which is about 5.24cm of steel surface length. There are between 5 and 26 laser samples taken per steel sample with the distribution shown in table \ref{tab:laser-sample-measurements}.

\begin{table}[htbp]
\centering
\caption{Number of Samples Grouped by Number of Measurements}
\label{tab:laser-sample-measurements}
\begin{tabular}{cc}
\toprule
Number of Samples & Number of Measurements \\
\midrule
4 & 5 \\
24 & 10 \\
1 & 23 \\
2 & 24 \\
13 & 25 \\
3 & 26 \\
\hline
\end{tabular}
\end{table}

It can be seen in Fig.~\ref{fig:zoomed_in_raw_sorm} \& \ref{fig:sorm_vs_our_Ra_calc_worst_outlier_combined} that there are many regions where the intensity is very low, indistinguishable from noise. We suspect this is due to the 1D nature of the measurement array, the laser will be reflected in a hemisphere but we are only collecting an arc.
The missing data might be one reason contributing to the inaccuracy of the baseline technique.

The conventional stylus measurements involve running a stylus along the surface of the steel. The stylus moves up and down as it navigates the topography and records the heights as a surface profile. From a surface profile, the roughness profile can be calculated by removing the longer wavelengths. The Ra parameter is calculated from the roughness profile. For our experiments, we aim to predict the Ra parameter without closed-form calculation using the equation.

\begin{figure}[ht]
\centering
\includegraphics[width=\linewidth]{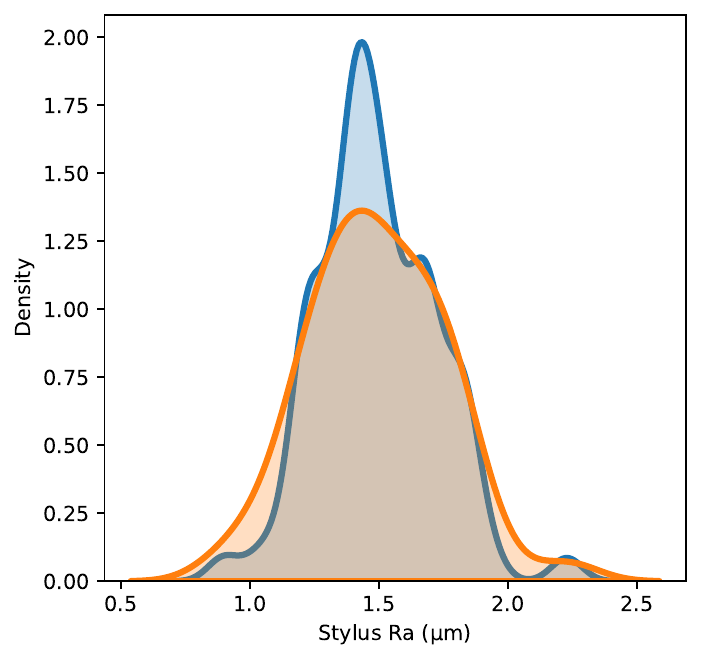}
\caption{The distribution of Ra for the steel samples in our dataset. Blue shows the distribution for all stylus measurements across all of the steel samples. Orange shows the distribution of the mean Ra measurements, one for each steel sample.}

\label{fig:dist_of_ra}
\end{figure}

The raw laser reflection data is the input for our experiments and the roughness parameter Ra calculated from the stylus profiles are our true outputs.
The distribution of Ra in our dataset can be seen in Fig.~\ref{fig:dist_of_ra}.

\subsection{Many-to-Many Data Problem}\label{sec:data:many_to_many}

\begin{figure}[ht]
\centering
\includegraphics[width=\linewidth]{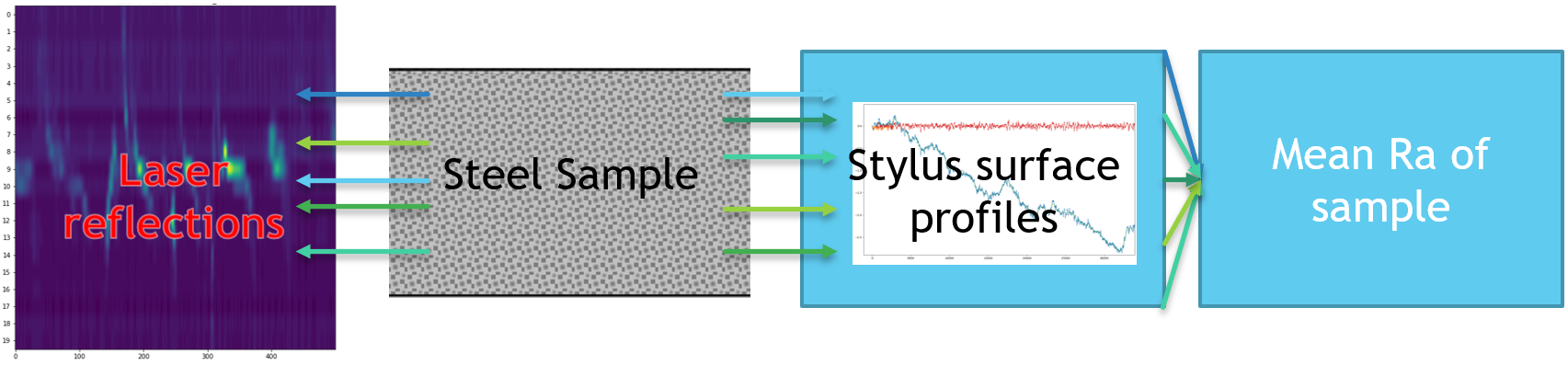}
\caption{The relation between the laser reflection data and the mean Ra of a steel sample. 
} 

\label{fig:many_to_many}
\end{figure}

Ideally, there would be a one-to-one map between a laser measurement and a stylus measurement, such that both measure the same track on the surface.
However, it is not feasible to gather this data as we do not have the capability to run the stylus on the same track as the laser due to the small scale and level of precision that would require. There is also the issue of the stylus head having a different size to that of the laser beam and coming into contact with a different area of steel. 

Therefore the data has an issue where for each of the 49 steel samples there are many laser reflection measurements and many stylus measurements all with different tracks on the surface without local spatial relation.

As a result, the mean Ra measurement from all stylus measurements for the steel sample must be calculated and used as the Ra of that steel sample. During training, all of the laser reflection samples from each steel sample will have the mean Ra of the steel sample be the label.

We expect that this will result in additional noise in our model, but expect the model performance to still be accurate, as steel samples naturally have variation across the same sample.

\section{Methodology}\label{sec:methodology}

We apply deep learning models from different areas of research in order to judge their suitability for the problem in terms of accuracy and speed. This will be compared to the baseline of calculating an Ra from a profile built from gradients. As the data is 2-dimensional data, we include deep learning models commonly used for image classification. Classification models can be modified to produce a regression output by changing the output layer of the model. We also experiment with the top 1-dimensional deep learning time series classification models, treating each of the individual sensors as a separate channel. We also note that these models are designed with data that has a long time step dimension in mind, whereas the 2-dimensional models typically come from machine vision and expect a square input image with a substantially shorter length dimension than ours of 65536. 1-dimensional models also have the potential to be faster due to lower computational complexity.

\subsection{Baseline calculation}

To calculate the baseline, the gradient profile first needs to be calculated.

\begin{figure}[ht]
  \centering
  \begin{subfigure}[b]{\linewidth}
    \centering
    \includegraphics[width=\linewidth]{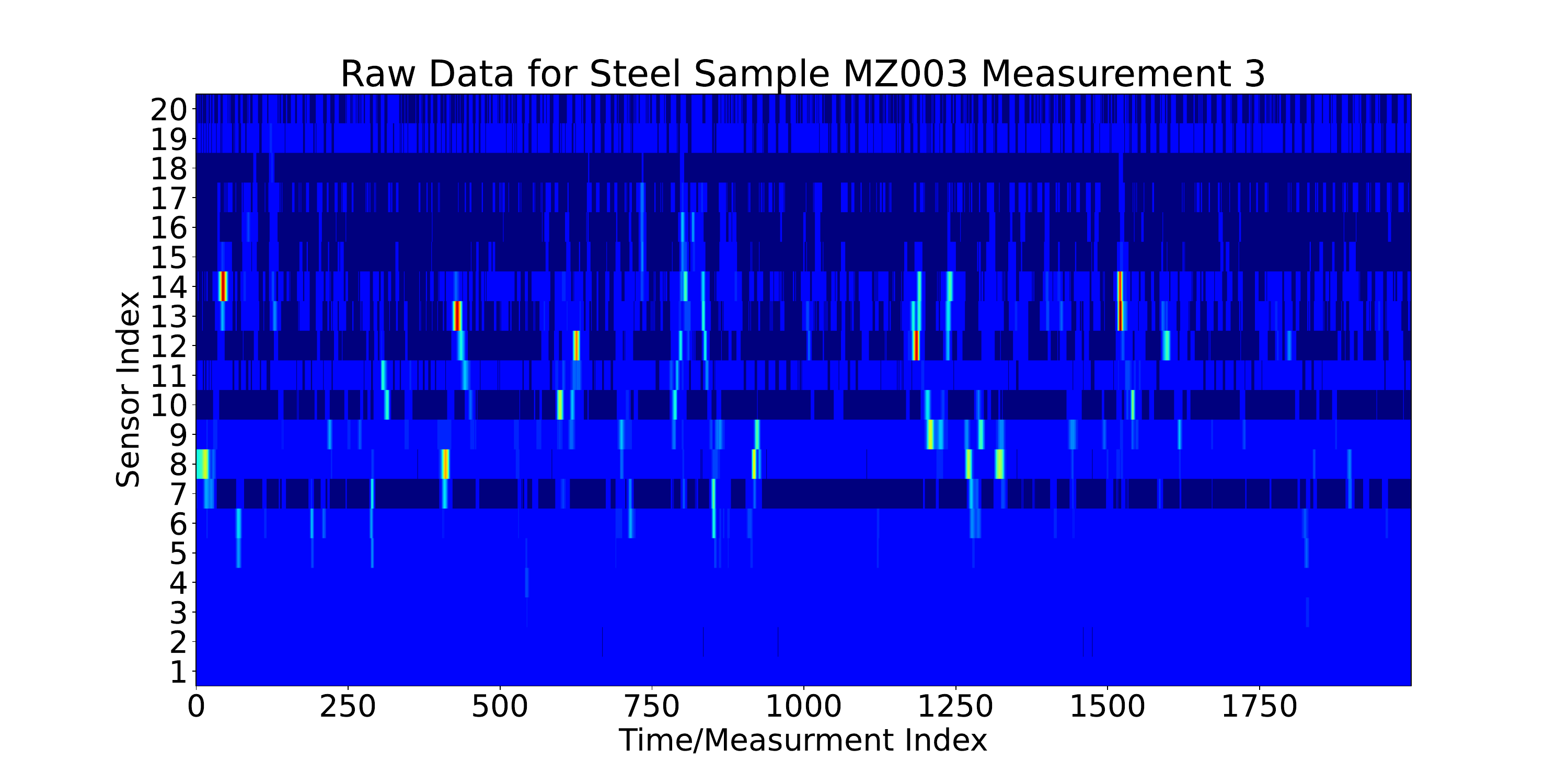}
    \caption{No Thresholding}
    \label{fig:sorm_vs_our_Ra_calc_worst_outlier_no_thresh}
  \end{subfigure}
  
  \begin{subfigure}[b]{\linewidth}
    \centering
    \includegraphics[width=\linewidth]{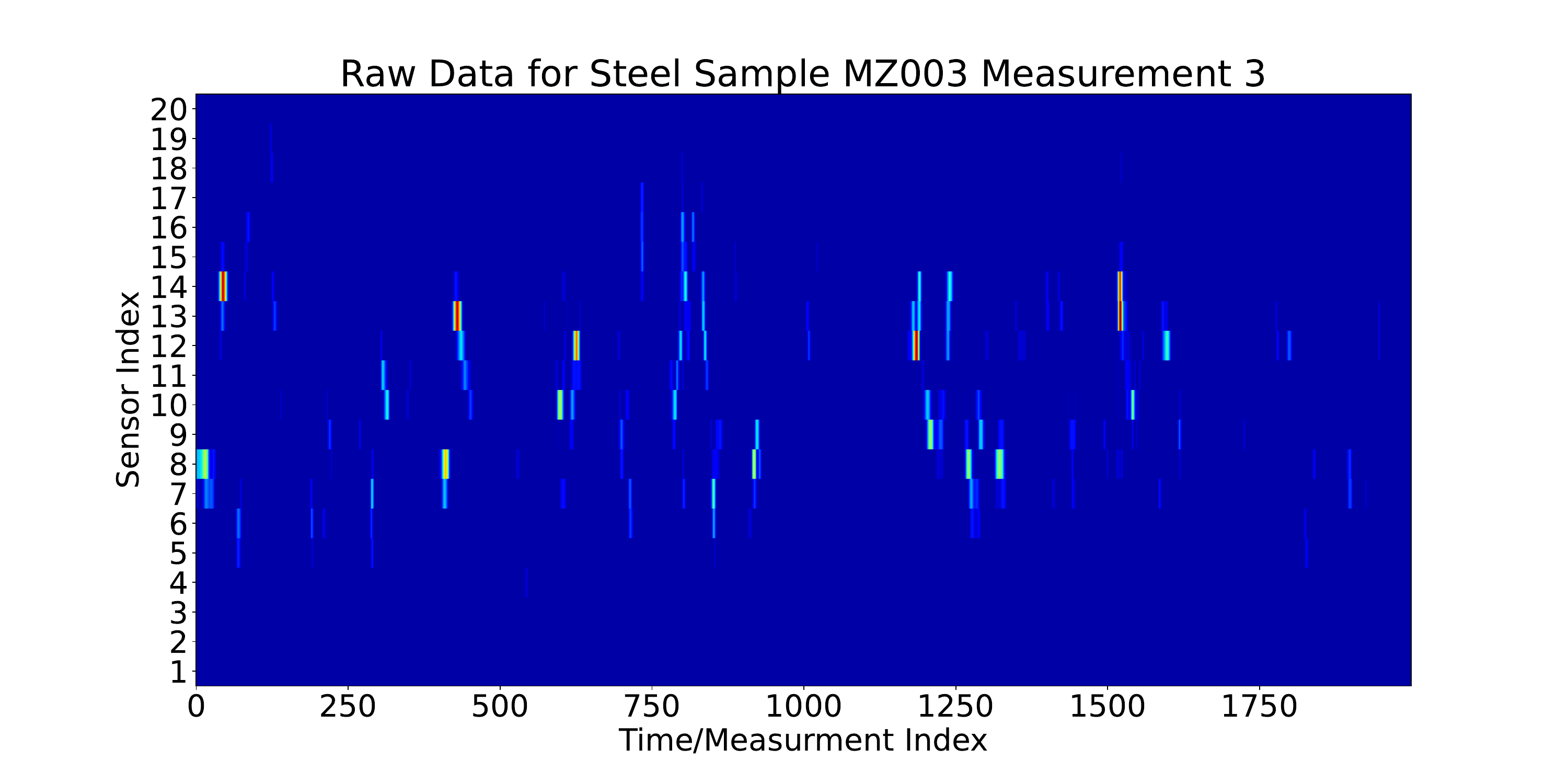}
    \caption{With Thresholding}
    \label{fig:sorm_vs_our_Ra_calc_worst_outlier}
  \end{subfigure}
  
  \caption{Comparison of raw intensity data for the worst outlier with and without thresholding.}
  \label{fig:sorm_vs_our_Ra_calc_worst_outlier_combined}
\end{figure}

A single laser intensity reading \( X \) is a matrix with shape \( 20 \times 2^{16} \) denoted as \( X_{i,j} \), where \( i \) represents the sensor index and \( j \) represents the timestep index. 
First perform a thresholding operation expressed as:

\begin{equation}
\tilde{X}_{i,j} = (X_{i,j} - \text{min}_{\theta}(X_{i,:}))_{+}
\label{eq:threshold}
\end{equation}

\noindent where $\text{min}_{\theta}(X_{i,:})$ orders $X_{i,:}$ and takes the value at position $\theta$. For example, if the threshold value $\theta$ is $2$ as used in for data, then $\text{min}_{\theta}(X_{i,j})$ takes the 2nd smallest value of $X_{i,:}$ for the channel $i$. The notation $(\cdot)_{+}$ denotes the function that maps any negative input value to zero and leaves all non-negative values unchanged. Therefore, the $\theta$ smallest values in each channel of $X$ are set to zero and the other values are shifted accordingly. An example of thresholded and non-thresholded data can be seen in Fig. \ref{fig:sorm_vs_our_Ra_calc_worst_outlier_combined}.

For calculating the gradients, denoted $g$, a vector $A$ is used. It has length 20 denoted as \( A_i \), where \( i \) represents the angle corresponding to the sensor at that index. 
Then, for each time step \( j \), the operation can be expressed as:

\begin{equation}
g_j = \frac{1}{2} \cdot \frac{\sum_{i=1}^{20} X_{i,j} \cdot A_i}{\sum_{i=1}^{20} X_{i,j}}
\label{eq:grads}
\end{equation}


\noindent where \(g_j \) represents the value of the resulting vector at time step \( j \). Note that the summation is performed over the 20 channels in matrix \( X \), and the division is by the sum of the values in matrix \( X \) for the corresponding time step \( j \). Dividing by the sum of the intensities makes the resulting gradient the angle corresponding to the mean intensity, and multiplying by \( \frac{1}{2} \) ensures that the surface gradients, rather than the reflected gradients are obtained.

Due to certain time steps where the sensors do not detect any light, some of the gradient values become infinite. To fill in these missing values, linear interpolation is used.

To obtain the surface profile from the calculated gradients, you accumulate the tangent of the gradient values along the desired direction over discrete time steps. The integrated surface profile, denoted as $ \text{surface}_i $, can be calculated with:

\begin{equation}
\text{surface}_i = \sum_{j=0}^{i} \tan(g_j) \Delta t
\label{eq:sum_grads}
\end{equation}



\noindent where $ i $ represents the discrete time step index, $ \Delta t $ represents the time step increment of $0.8$, and $ \text{surface}_i $ represents the accumulated surface profile at the $ i $-th time step. 

To calculate the roughness profile the longer wavelengths are filtered out from the signal. 
Extend the surface profile by attaching flipped profiles to both ends. This is done by concatenating the flipped profile of length \(N\) to the beginning and end of the original profile, resulting in an extended profile of length \(3N\). 

Then perform the Fast Fourier Transform (FFT) on the extended surface profile, which gives the profile in the frequency domain, denoted as \(F(\omega)\), where \(\omega\) is the frequency coordinate. A frequency-domain filter is next applied to \(F(\omega)\) to remove the longer wavelengths that correspond to the waviness component. This can be done by setting the Fourier coefficients of the unwanted frequencies to zero. The cutoff frequency can be determined based on the desired waviness wavelength, in our case \SI{80}{\micro\meter}. 

The inverse FFT of the filtered data obtains the roughness profile in the spatial domain, denoted as \(R(x)\), where x is the index. 
The extended roughness profile needs to be cropped to the length of the original profile, \(R(x)\), where \(x > N\) and \(x < 2N\).

\begin{figure}[ht]
\centering
\includegraphics[width=\linewidth]{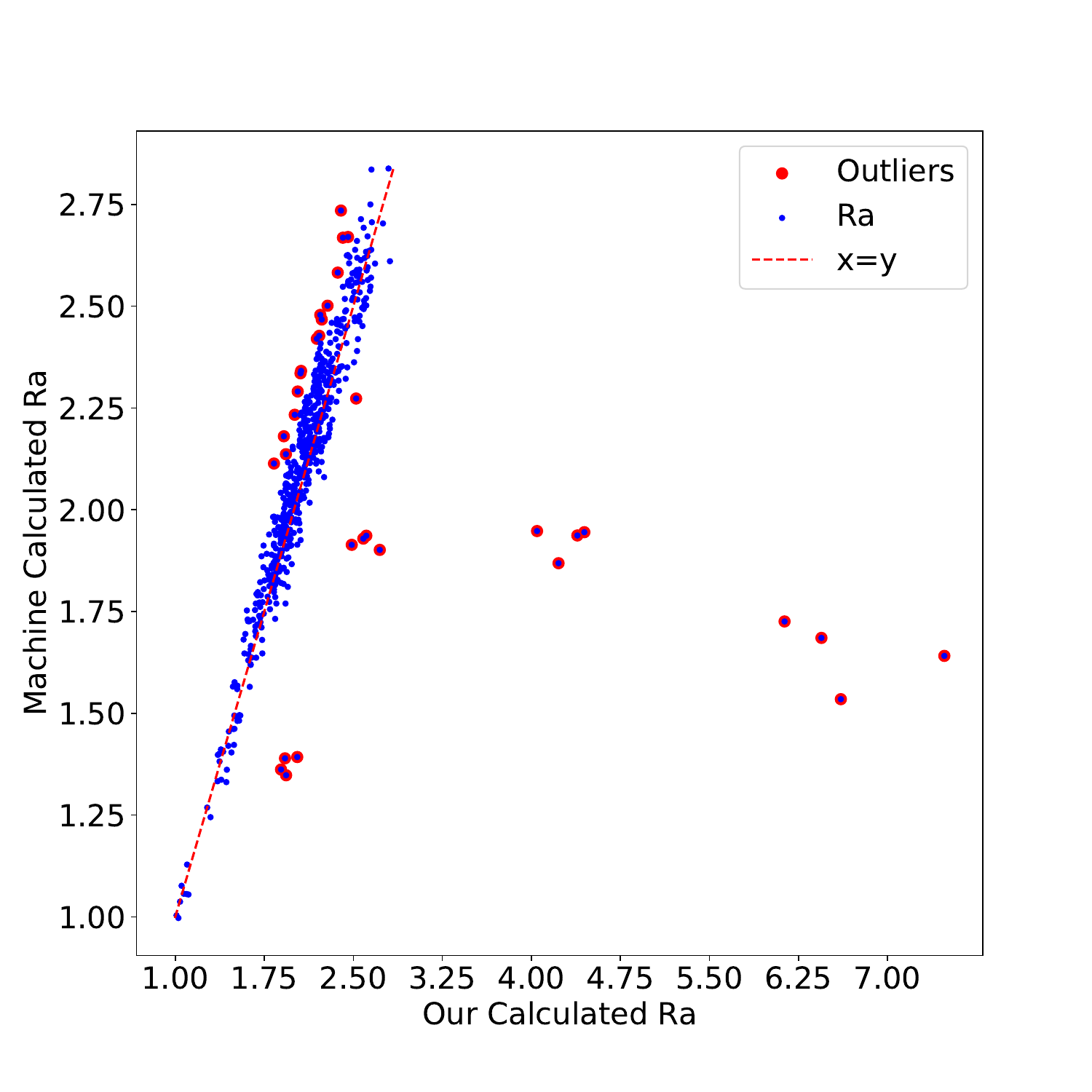}
\caption{The difference between the Ra calculated from the raw laser-reflected data by us and by the measurement device itself.}

\label{fig:sorm_vs_our_Ra_calc}
\end{figure}

The resulting roughness profile represents the filtered surface profile data with the waviness component removed, leaving only the roughness component. Apply Eq. \ref{eq:Ra} to the roughness profile to calculate the Ra parameter. 


The laser measurement device provides outputs of the raw laser reflection intensities through time, but also an Ra value it has calculated itself. Fig.~\ref{fig:sorm_vs_our_Ra_calc} shows the difference between the values calculated from the raw input vs. the values calculated by the measurement device from the raw input. The plot shows that most results are very similar, but that there are some very large outliers. 

A plot of raw intensity data can be seen in Fig.~\ref{fig:sorm_vs_our_Ra_calc_worst_outlier_combined} in raw unprocessed form, and with thresholding to remove sensor issues at very low intensities. The worst outliers have lots of gaps in the intensity data which is likely what is causing the large discrepancies in our calculations. We believe that the measurement device might have a better method for interpolating the raw data than the one we have used. It is not in the scope of our work to re-engineer the closed-form solution exactly as the measurement device calculates it, we just provide here an insight into the process. Therefore, we use the values from the measurement device as the closed-form baseline values and not our calculated values.

\subsection{Models tested}\label{sec:methodology:models_tested}



In this section, we present the data-driven models that we test in our study, categorized into three groups: non-deep learning, 1D deep learning, and 2D deep learning.

For the 1D models, we selected models from the Time Series Extrinsic Regression (TSER)~\cite{tan2021TSER} and Time Series Classification (TSC)~\cite{ismail2019deepTSC} areas of research. These models are specifically designed for data with a long length, which aligns with our data characteristics. In our approach, we treat the data from each of the 20 sensors as individual data channels.

For 2D deep learning, we explored models from the machine vision field, considering that our data has a 2D structure, where the channels are spatially related to each other and the time dimension represents a scan of the surface length. It is important to note that machine vision models are typically used on square-shaped images, while our data is thin and long. To adapt these models to our data, we treated it as a single-channel grayscale image with a height of 20 (representing the sensors) and a width of $2^{16}$.

For the implementation of the 1D deep learning models, we utilized the tsai package~\cite{tsai}, which offers convenient implementations of many of these models. Meanwhile, we modified several 2D models from the TorchVision package~\cite{torchvision2016} to suit our specific data format and experimental requirements, and use the fastai package for the xresnet models~\cite{howard2018fastai}. 

\subsubsection{Non-Deep Learning (Data-Driven Approach)}

Rocket, initially proposed by Angus Dempster et al. \cite{dempster2020rocket} as an ``exceptionally fast and accurate" model for TSC, stands out amongst other state-of-the-art models in its field due to its fast training speed. Rocket's feature extraction process yields broad and informative features, making it adaptable for regression tasks, which can be achieved by replacing the ridge classifier with a ridge regressor. This adaptability has been demonstrated by Chang Wei Tan et al. \cite{tan2021TSER}, where Rocket outperformed other models and emerged as the best-performing model in their TSER experiments.

The architecture of Rocket is composed of 10,000 parallel convolutional operations, each with random parameterization. The convolutional output then undergoes two global pooling operations for each kernel, resulting in 20,000 features. These two pooling operations are max pooling and Proportion of Positive Values operation (PPV). The ridge classifier utilizes these features to make predictions. In the original paper, Rocket achieves high accuracy while exhibiting the shortest training phase when evaluated across the 85 UCR TSC datasets~\cite{dau2019UCRarchive}.

MiniRocket, introduced by Angus Dempster et al. \cite{dempster2021minirocket}, builds upon the efficiency of the ROCKET model, further distinguishing itself from other state-of-the-art approaches in terms of speed. It offers notable advancements in accuracy relative to computational cost. 
This efficiency boost is achieved through modifications to the kernel parameterization. By employing binary kernels with two potential values (-1 or 2), MiniRocket performs convolution operations based on addition rather than multiplication. With only two kernel values available, there exist 84 unique kernels. To approximate the use of 10,000 kernels, these 84 kernels have 119 dilations, chosen based on the input data length, resulting in 9,996 unique kernels. Additionally, Biases are sampled from the input data, resulting in a normalization effect. Furthermore, MiniRocket eliminates the max pooling operation, opting for only PPV pooling, yielding one feature per kernel, amounting to approximately 10,000 features compared to ROCKET's 20,000. The authors report that MiniRocket achieves processing speeds roughly 75 times faster on larger datasets while maintaining similar accuracy to ROCKET, a model already recognized for its efficiency compared to other models in the existing literature.


XGBoost, introduced by Chen and Guestrin \cite{chen2016xgboost}, is a highly effective and widely used machine learning algorithm. It combines gradient boosting and tree-based models to handle diverse data types and capture complex interactions. XGBoost sequentially adds decision trees to a growing ensemble, where each subsequent tree corrects the errors made by the previous trees. Additionally, XGBoost supports various objective and loss functions and employs regularization techniques such as shrinkage and feature subsampling, which help prevent overfitting and enhance generalization. Its exceptional performance, versatility, and scalability have made it a popular choice in competitions and real-world applications. 

Random Forest, introduced by Breiman \cite{breiman2001random}, is a powerful ensemble learning algorithm that combines multiple decision trees for accurate predictions. It handles high-dimensional data, reduces overfitting, and provides insights into feature importance. With its versatility, efficiency, and interpretability, Random Forest is widely used in various domains and applications.
The algorithm works by constructing a multitude of decision trees on randomly selected subsets of the training data. Each tree in the Random Forest operates independently, making predictions based on the majority vote of the individual trees. This ensemble approach helps to reduce overfitting and improves the overall robustness of the model. Random Forest has demonstrated its effectiveness across various domains and applications, including classification, regression, and feature selection.

\subsubsection{1D Deep Learning (Data-Driven Approach)}


Bai et. al. propose a network that uses dilated causal convolutions to capture temporal dependencies~\cite{bai2018TCN}, the Temporal Convolutional Network (TCN). The TCN model uses stacked convolutional blocks with each block having exponentially increasing dilations, d. This helps capture long-range patterns such that the model has a large receptive field by the final layer.  
The Receptive Field (RF) of the model can be estimated with the following equation:
\begin{equation}
RF = 2 \cdot \sum_{i=0}^{n-1} (k_s - 1) \cdot d^i
\label{eq:tcn_receptive_field}
\end{equation}
Each convolutional block contains two convolutional, hence the multiplication by two in the equation. 
The model has two additional hyperparameters, kernel size (ks or $k_s$ in the equation) and number of layers (nl or n in the equation). These can be selected to increase model complexity and to change the receptive field, as it is likely beneficial for the receptive field to encompass the data length.

The model makes use of residual connections between blocks, weight norm before convolution, and Dropout in the network.  We use the tsai Python package~\cite{tsai} for the implementation of the model in PyTorch. Three different configurations of kernel size (ks) and number of layers (nl) are explored: ks=9, nl=12; ks=5, nl=13; and ks=7, nl=8. These configurations yield approximate receptive fields of 65520, 65528, and 3060 respectively. It is worth noting our data length, which is 65536. 

InceptionTime~\cite{ismail2020inceptiontime} is a deep learning architecture inspired by the Inception module~\cite{szegedy2017inceptionV4} originally proposed by Fawaz et. al. for TSC. 
It incorporates blocks of different-sized convolutions, accompanied by residue connections, to effectively extract both local and global features from the input data. Each block applies a bottleneck layer to the input, which reduces the dimensionality of the data. Subsequently, it employs 10, 20, and 40-length convolutional filters within the blocks. Additionally, an alternative path applies a max pooling layer to the input, followed by a bottleneck layer. To obtain a global representation of the features, a Global Average Pooling (GAP) layer is employed. Finally, the architecture concludes with a softmax classification layer for accurate predictions. As we are performing regression, we use a linear prediction layer for the output layer without softmax.

Rahimian et al. propose XceptionTime~\cite{rahimian2019xceptiontime} as a dedicated model for analyzing surface Electromyography (sEMG) signal data. The authors draw inspiration from the success of two existing models: Xception, known for its superior performance in large image classification compared to Inception V3, and InceptionTime, which was developed based on the Inception V4 concept. Motivated by these achievements, the authors introduce XceptionTime as a novel model that combines the strengths of InceptionTime with the utilization of depthwise separable convolution Xception modules. By incorporating depthwise separable convolutions, XceptionTime outperforms its existing counterparts in analyzing sEMG signal data.

The MRNN-FCN model architecture, proposed by Khan et. al.~\cite{khan2020MRNN}, incorporates two parallel feature extraction routes, which are subsequently concatenated in the final feature layer. Path 1, begins with a dimension shuffle operation, followed by the selected Recurrent Neural Network (RNN) unit and dropout regularization for training purposes.

Path 2 encompasses a combination of different layers. It begins with a 1D convolutional layer with Batch normalization and ReLU activation. Subsequently, a squeeze and excitation block is employed.
Additional convolutional layers and squeeze and excitation blocks follow. Finally, a GAP operation is used to obtain a summary representation of the extracted features.

The outputs from Path 1 and Path 2 are concatenated, facilitating the fusion of diverse features derived from different pathways, and providing a comprehensive representation of the input data.

The MRNN-FCN, MGRU-FCN, and MLSTM-FCN models share a similar architecture, with the key difference lying in the type of RNN module utilized. While the MRNN-FCN employs a standard RNN module, the MGRU-FCN and MLSTM-FCN models replace it with the Gated Recurrent Unit (GRU) and Long Short-Term Memory (LSTM) modules, respectively. The GRU and LSTM modules are variants of the RNN module that address the vanishing gradient problem and enable the model to capture long-term dependencies more effectively.

He et al. \cite{resnet_original} introduced the ResNet framework to facilitate the training of significantly deeper networks compared to previous approaches. This framework has revolutionized the fields of computer vision and deep learning, as residual learning and skip connections have become integral components in many modern deep architectures. Residual learning, in combination with skip connections, represents the key innovation of ResNet. It enables the network to bypass multiple layers and directly propagate information from earlier layers to subsequent ones, thereby ensuring a smooth flow of gradients during training and mitigating the vanishing gradient problem.

The ResNet framework is structured as a sequence of layers, with each layer comprising residual blocks. These blocks consist of multiple convolutional layers, batch normalization, Rectified Linear Unit (ReLU) activation, and a shortcut connection that adds the input of the block to its output. 
The architecture is designed to accommodate deeper networks, typically achieved by increasing the number of channels while reducing spatial dimensions. This is accomplished by downsampling and progressively augmenting the number of channels in each subsequent layer. This architectural strategy empowers ResNet models to effectively capture and process complex features at varying scales and depths, leading to enhanced performance in diverse computer vision tasks.

In our experiments, we have utilized the ResNet model proposed by Wang et al. \cite{TSC_from_scratch_dl} specifically designed for 1D time series data. This implementation deviates from conventional ResNet architectures in several aspects. Firstly, it employs 1D convolutions instead of 2D convolutions, thereby accommodating the nature of time series data. Unlike traditional ResNet models, this variant does not incorporate downsampling operations. However, it still increases the number of channels across layers to capture diverse temporal patterns. Moreover, this model diverges in terms of its depth, comprising only three layers as opposed to the standard ResNet architecture with four layers. Furthermore, it introduces a decreasing kernel size pattern across the three layers, transitioning from a kernel size of 7, to 5, and finally to 3. Typically, ResNet architectures employ kernel sizes of 3 throughout all layers, except for the input layer.

Wang et al. \cite{TSC_from_scratch_dl} also propose an additional model, the Fully Convolutional Network (FCN).
The FCN is a straightforward model composed of three consecutive 1D convolutional blocks. These blocks extract feature channels in a sequence of 128, 256, and 128, respectively. Following each convolutional block, batch normalization and ReLU activation functions are applied. To obtain a global representation, global average pooling is employed. The final prediction is made using a linear layer.

He et al. \cite{he2019xResbagOfTricks} proposed a series of modifications, referred to as a ``bag of tricks", that can be applied to the ResNet architecture, resulting in variants known as xresnets. While xresnets maintain the core principles of ResNet, utilizing residual blocks with convolutional layers and shortcut connections, xresnets incorporate notable structural changes.
In particular, xresnets modify the input stem by splitting the original convolutional layer with a kernel size of 7 and a stride of 2 into three separate convolutional layers. Each of these layers utilizes a kernel size of 3. The first layer employs a stride of 2, while the subsequent two layers have a stride of 1.
Additionally, xresnets modify the downsampling of the identity path. Instead of employing a kernel size of 1 with a stride of 2, xresnets utilize an average pooling layer with a kernel size of 2, followed by a kernel size 1 convolutional layer with step size of 1. 
Furthermore, xresnets introduce an expansion parameter, which, when not a value of 1, adds an additional convolutional layer to each block and augments the number of channels. 
We use the 1D variants, xresnet1d18, xresnet1d34, xresnet1d50, xresnet1d101, xresnet1d18 deep, xresnet1d34 deep, xresnet1d50 deep, xresnet1d18 deeper, xresnet1d34 deeper, and xresnet1d50 deeper.

The ResCNN network proposed by Zou et. al.~\cite{ResCNN} is composed of four sequential blocks. The first block features two paths: the first path includes three convolutional blocks, each incorporating batch normalization and ReLU activation, while the second path consists of a single convolutional block with batch normalization and no ReLU activation. The output of the second path acts as a shortcut and is added to the output of the first path, followed by a ReLU activation. 
Subsequently, each block performs 1D convolution, followed by batch normalization and an activation function. The first block employs leaky ReLU activation, the second block utilizes Parametric ReLU (PReLU) activation, and the final block applies Exponential Linear Unit (ELU) activation.
To capture a global representation of the feature maps, global average pooling is performed. Finally, a linear layer is used for prediction.

\subsubsection{2D Deep Learning (Data-Driven Approach)}

For the 2D models selected, we have been more selective due to the increased computation required to experiment with these models. Two variations of xresnet, xresnet18, and xresnet34 have been chosen. These two models are the same as previously described in the 1D section but instead with 2D filters being used.

ConvNeXt, introduced by Liu et al.~\cite{liu2022convnext} in the era dominated by transformer architectures, offers a compelling alternative to traditional ConvNets. While ViTs quickly surpassed ConvNets as the state-of-the-art image classification model, they encountered challenges when applied to general computer vision tasks. Hierarchical Transformers, like Swin Transformers, reintroduced ConvNet priors, making Transformers more applicable to various vision tasks. The authors incorporate design elements from Vision Transformers into a standard ResNet architecture.  unleash the potential of pure ConvNets. The result is ConvNeXt, a family of models constructed entirely from standard ConvNet modules. ConvNeXt achieves remarkable accuracy and scalability, outperforming Swin Transformers on multiple benchmarks, while retaining the simplicity and efficiency of ConvNets. These findings challenge prevailing beliefs and emphasize the importance of convolution in computer vision. Two variants are used: Tiny and Small.

The Vision Transformer (ViT), introduced by Dosovitskiy et al.~\cite{dosovitskiy2020vit}, represents a significant breakthrough in computer vision by leveraging transformer-based architectures, which have excelled in natural language tasks. ViT challenges the conventional wisdom that Convolutional Neural Networks (CNNs) are not necessary and a pure transformer can perform very well for image classification. The self-attention mechanisms of transformers, applied to embedded patches, capture global contextual information from input images, enabling the model to comprehend intricate visual patterns and relationships. The self-attention mechanism facilitates efficient computation of pairwise interactions between image patches, allowing ViT to model long-range dependencies effectively. 
ViT directly applies a pure transformer encoder architecture to sequences of embedded image patches.
The ViT architecture consists of an initial patch embedding layer, which splits the input image into a sequence of fixed-size patches. These patches are then linearly projected into a set of learnable embeddings. The resulting embeddings are augmented with positional encodings to encode spatial information. The transformer encoder layers process the embedded patches, incorporating self-attention and feed-forward neural network modules. The self-attention mechanism captures global context by attending to all image patches, enabling effective information exchange. Finally, a classification token is added, and the resulting sequence is passed through a linear layer followed by softmax activation to obtain the class probabilities. We have changed the model slightly by adjusting the patch size to 20 by 32 instead of the 16 by 16 patch size in the original model, such that the patches fit the 20 sensor dim exactly and the 32 ($2^{5}$) is divisible by the $2^{16}$ length of the data, resulting in 2,048 patches. 
We experiment with the ViT Tiny and the ViT Small models.

Swin v2, introduced by Liu et al.~\cite{liu2022swinv2}, improves upon previous transformer-based models which perform global image processing. It adopts a hierarchical processing strategy to effectively handle high-resolution images. Building upon the Swin Transformer architecture, which reintroduces ConvNet priors, Swin v2 serves as a practical choice for a generic vision backbone. By dividing the input image into non-overlapping patches and organizing them into stages and windows, the model captures both local and global information efficiently. Additional enhancements, such as the patch merging module and shift operation, contribute to improved feature extraction and contextual modeling. In our experiments, we evaluate the performance of the Swin v2 Tiny and Swin v2 Small variants.

These models were chosen based on their proven performance in various machine learning tasks and their potential to improve the accuracy of laser reflection measurements for online sheet steel manufacturing.

\subsection{Training and Hyperparameters}\label{sec:training_deets}

For each deep learning experiment, the PyTorch framework is employed, utilizing a consistent training loop and set of hyperparameters. The chosen optimizer is Lion~\cite{chen2023symbolic}, with a weight decay of 0.01. To assess the performance, the Mean Squared Error (MSE) loss function is used, as defined in Eq.~\ref{eq:MSE}. The initial learning rate is set to 0.0001, and a learning rate scheduler that based on the reduce-on-plateau strategy is utilized. This scheduler reduces the learning rate by a factor of 0.5 when the validation loss fails to improve within the cooldown and patience intervals. In our experiments, the cooldown and patience is set to 10 epochs. Training continues until the validation loss fails to improve for 100 consecutive epochs. Throughout the training process, the checkpoints are saved at regular intervals and the checkpoint weights with the best validation loss are selected by the model upon completion.

\begin{equation}
MSE = \frac{1}{n} \sum_{i=1}^{n}(y_i - \hat{y_i})^2
\label{eq:MSE}
\end{equation}

\subsection{Dataset preparation for training}\label{sec:methodology:data_prep}

Two methods of training are performed. In the first, 20\% of the data is held back for the test set in order to evaluate the models' performance. Due to the limited spread of the Ra label values from the many-to-many problem, we perform the split on each steel sample in order to guarantee that the model has seen 80\% of the input samples from each steel sample. 

In the second method, due to the relatively small dataset size, with only 49 different steel samples, we believe that it is important to test the model's generalization ability on completely unseen new steel samples. This is done by sectioning our data into k folds, where each of the folds contains all measurements from a single steel sample. A new model is trained for each fold where one of the folds is used as the test set and all the other folds as training data. 

As the model has not seen this steel sample before, these experiments ensure the model is not simply grouping data samples assigning the related Ra from the same steel samples and instead is actually learning a useful transformation from Raw reflected readings to the Ra.


Before feeding the data into the model, thresholding is performed as specified in Eq. \ref{eq:threshold} and then z-score normalization. Calculated mean and standard deviation values from the training data are used to normalize the test data.

\section{Results}\label{sec:results}

\begin{figure}[ht]
\centering
\includegraphics[width=\linewidth]{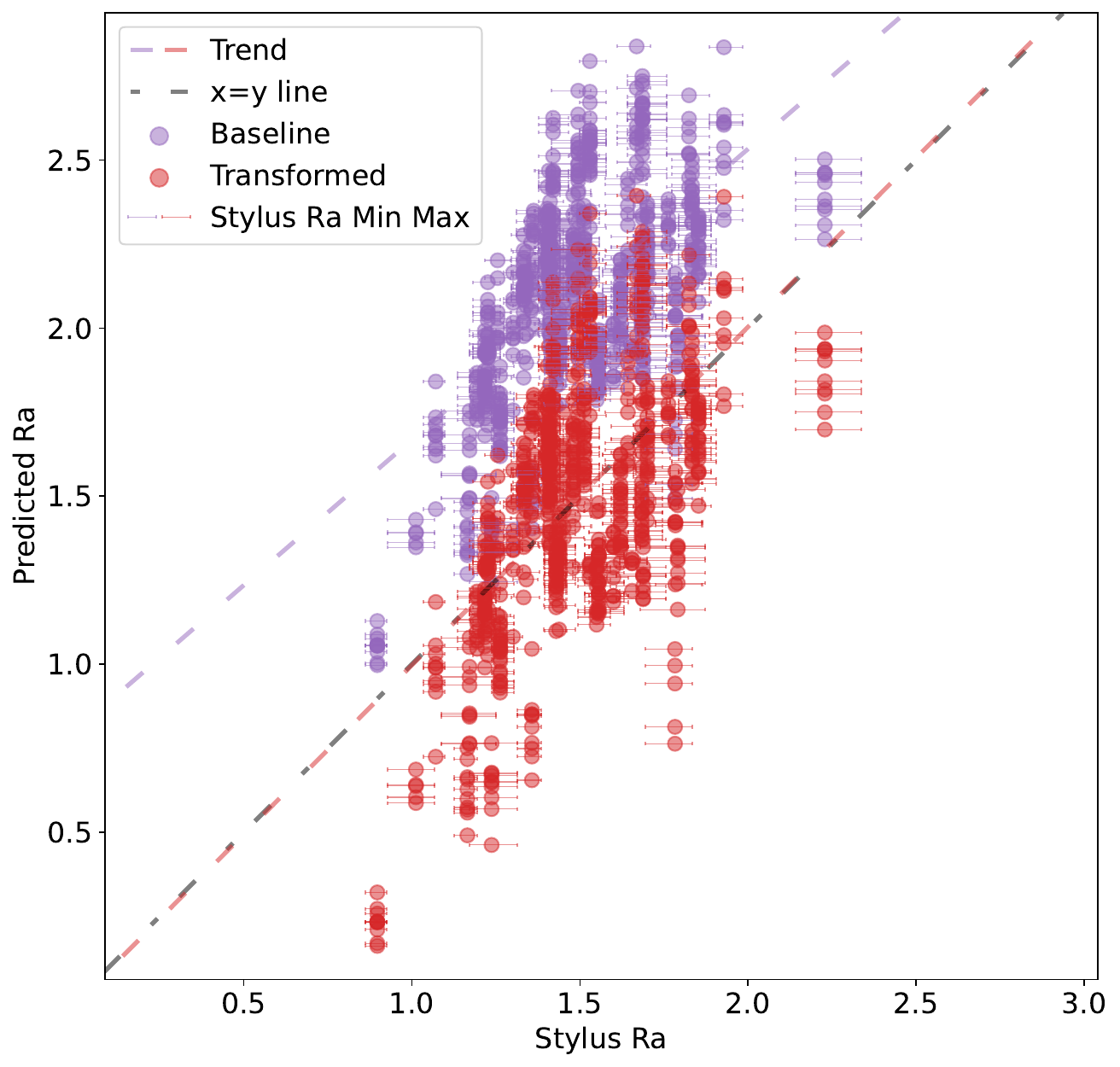}
\caption{The Closed-Form Baseline Ra results and the results transformed onto the $x=y$ line.}

\label{fig:baseline_scatter}
\end{figure}

We compare our data-driven approaches to the closed-form baseline for the calculation of the Ra from the laser reflection data. Fig.~\ref{fig:baseline_scatter} shows that in general, the baseline has a tendency to produce a higher Ra than the stylus. In order to improve the baseline results we have transformed the calculated Ra values onto the true x=y line. 

\subsection{20\% data split experiments}

\begin{table*}[ht]
\centering
\caption{Results on Data: 20\% per Sample}
\begin{tabular}{|c|c|c|c|c|}
\hline
\textbf{Model} & \textbf{RMSE} & \textbf{Correlation} & \textbf{Max Error} & \textbf{Pred. Coverage} \\
\hline
\multicolumn{5}{|c|}{\textbf{Closed-Form Baseline}}\\
\hline
\textbf{Baseline Transformed} & 0.3076 & 0.5940 & 1.0193 & 0.0992\\
Baseline & 0.6546 & 0.5940 & 1.2648 & 0.0064\\
\hline
\multicolumn{5}{|c|}{\textbf{Non-Deep Learning (Data-Driven Approach)}} \\
\hline
\textbf{MiniRocket} & 0.0660 & 0.9576 & 0.2078 & 0.5921\\
Rocket & 0.0882 & 0.9259 & 0.3876 & 0.4737\\
XGBoost & 0.1970 & 0.5431 & 0.7598 & 0.2303\\
Random Forest & 0.1985 & 0.6012 & 0.7038 & 0.2105\\
\hline
\multicolumn{5}{|c|}{\textbf{1D Deep Learning (Data-Driven Approach)}} \\
\hline
\textbf{TCN ks=9 nl=12} & 0.0487 & 0.9770 & 0.1740 & 0.6908\\
TCN ks=5 nl=13 & 0.0499 & 0.9762 & 0.1881 & 0.6776\\
TCN ks=7 nl=8 & 0.0529 & 0.9729 & 0.1891 & 0.6579\\
MGRU-FCN & 0.0564 & 0.9695 & 0.1948 & 0.6053\\
InceptionTime & 0.0584 & 0.9677 & 0.2476 & 0.6447\\
xresnet1d34 & 0.0618 & 0.9639 & 0.3486 & 0.5921\\
MLSTM-FCN & 0.0633 & 0.9609 & 0.2052 & 0.5789\\
xresnet1d18 & 0.0643 & 0.9597 & 0.3837 & 0.6382\\
xresnet1d50 deep & 0.0644 & 0.9600 & 0.3110 & 0.6118\\
xresnet1d101 & 0.0662 & 0.9587 & 0.2221 & 0.5329\\
MRNN-FCN & 0.0703 & 0.9517 & 0.3980 & 0.6513\\
xresnet1d50 deeper & 0.0704 & 0.9513 & 0.3141 & 0.5329\\
xresnet1d18 deep & 0.0712 & 0.9506 & 0.4461 & 0.6053\\
xresnet1d34 deep & 0.0743 & 0.9478 & 0.4354 & 0.5789\\
xresnet1d50 & 0.0754 & 0.9460 & 0.4752 & 0.5658\\
xresnet1d34 deeper & 0.0779 & 0.9405 & 0.3381 & 0.5461\\
xresnet1d18 deeper & 0.0795 & 0.9418 & 0.5058 & 0.5921\\
XceptionTime & 0.0896 & 0.9201 & 0.3476 & 0.4474\\
ResCNN & 0.0917 & 0.9190 & 0.5634 & 0.5987\\
FCN & 0.1123 & 0.8799 & 0.6776 & 0.5197\\
ResNet & 0.1223 & 0.8691 & 0.8352 & 0.6382\\
\hline
\multicolumn{5}{|c|}{\textbf{2D Deep Learning (Data-Driven Approach)}} \\
\hline
\textbf{xresnet34} & 0.0453 & 0.9806 & 0.1390 & 0.7237\\
ConvNeXt Small & 0.0554 & 0.9706 & 0.1982 & 0.6184\\
ConvNeXt Tiny & 0.0561 & 0.9701 & 0.1796 & 0.6382\\
xresnet18 & 0.0621 & 0.9629 & 0.2787 & 0.5658\\
Swin v2 Tiny & 0.1124 & 0.8737 & 0.4015 & 0.2961\\
Swin v2 Small & 0.1508 & 0.7551 & 0.8138 & 0.2961\\
ViT Small & 0.1729 & 0.6859 & 0.4881 & 0.2632\\
ViT Tiny & 0.1947 & 0.5813 & 0.5736 & 0.2171\\
\hline
\end{tabular}
\label{tab:20_pc_model_comparison}
\end{table*}

The first set of conducted experiments are on the data split such that 20\% of data samples from each steel sample are used as the test data. The results of the 20\% data split analysis are presented in Table~\ref{tab:20_pc_model_comparison}, which provides an overview of the performance metrics for the evaluated models, categorized based on their approach. The analysis includes three approach types: closed-form baselines, non-deep learning (data-driven approaches), and deep learning (data-driven approaches). The metrics used for evaluation include the Root Mean Squared Error (RMSE)-defined in Eq.~\ref{eq:RMSE}, Pearson's Correlation, maximum prediction error, and the percentage of predictions within the minimum and maximum values obtained from the stylus measurement for the steel sample. The table is sectioned by type of approach, where each section is in descending order of RMSE with the best result displayed in bold.

\begin{equation}
RMSE = \sqrt{\frac{1}{n} \sum_{i=1}^{n}(y_i - \hat{y_i})^2}
\label{eq:RMSE}
\end{equation}

\noindent The \textbf{Closed-Form Baseline} models, including Baseline Transformed and Baseline, served as a reference point for comparison. These models achieved RMSE values of 0.3076 and 0.6546 respectively. 
Notably, the prediction coverage was relatively low for both models, with values of 0.0992 and 0.0064, respectively, meaning that many results fell outside the range of values provided by the stylus measurement.

For the \textbf{Non-Deep Learning (Data-Driven Approach)} models, the MiniRocket model demonstrated the best performance, achieving the lowest RMSE value of 0.0660 and a high correlation coefficient of 0.9576. 
The Rocket model achieved a slightly lesser performance, with an RMSE of 0.0882
In contrast, the XGBoost and Random Forest models performed poorly compared to the other models tested, as well as in comparison to the baseline models.
For XGBoost and Random Forest models, the input data must be completely flattened, resulting in a very large feature space of 1,310,720 (20x2**16) dimensions for each data sample. We suspect that the results suffer from the large dimensions and would benefit from some dimensionality reduction, such as PCA before data is input into the model. 

In the category of \textbf{1D Deep Learning (Data-Driven Approach)}, the models demonstrated strong performance. The TCN ks=9 nl=12 model achieved the lowest RMSE of 0.0487, with a high correlation coefficient of 0.9770. 
Other models in this category, such as TCN ks=5 nl=13, TCN ks=7 nl=8, MGRU-FCN, and InceptionTime, demonstrated comparable performance with RMSE values ranging from 0.0499 to 0.0584. 
Additionally, models such as xresnet1d34, MLSTM-FCN, and xresnet1d50 deep exhibited good performance with RMSE values below 0.065 and correlation coefficients above 0.9500.
It is interesting that the TCN ks=7 nl=8 performed similarly to the other two models despite this variation not having a receptive field large enough to encompass the entire data length. This might be due to the parameter Ra being, in essence, a mean across the length of the sample, such that the output regression layer can easily combine the features from different regions.

\begin{figure}[t]
\centering
\includegraphics[width=\linewidth]{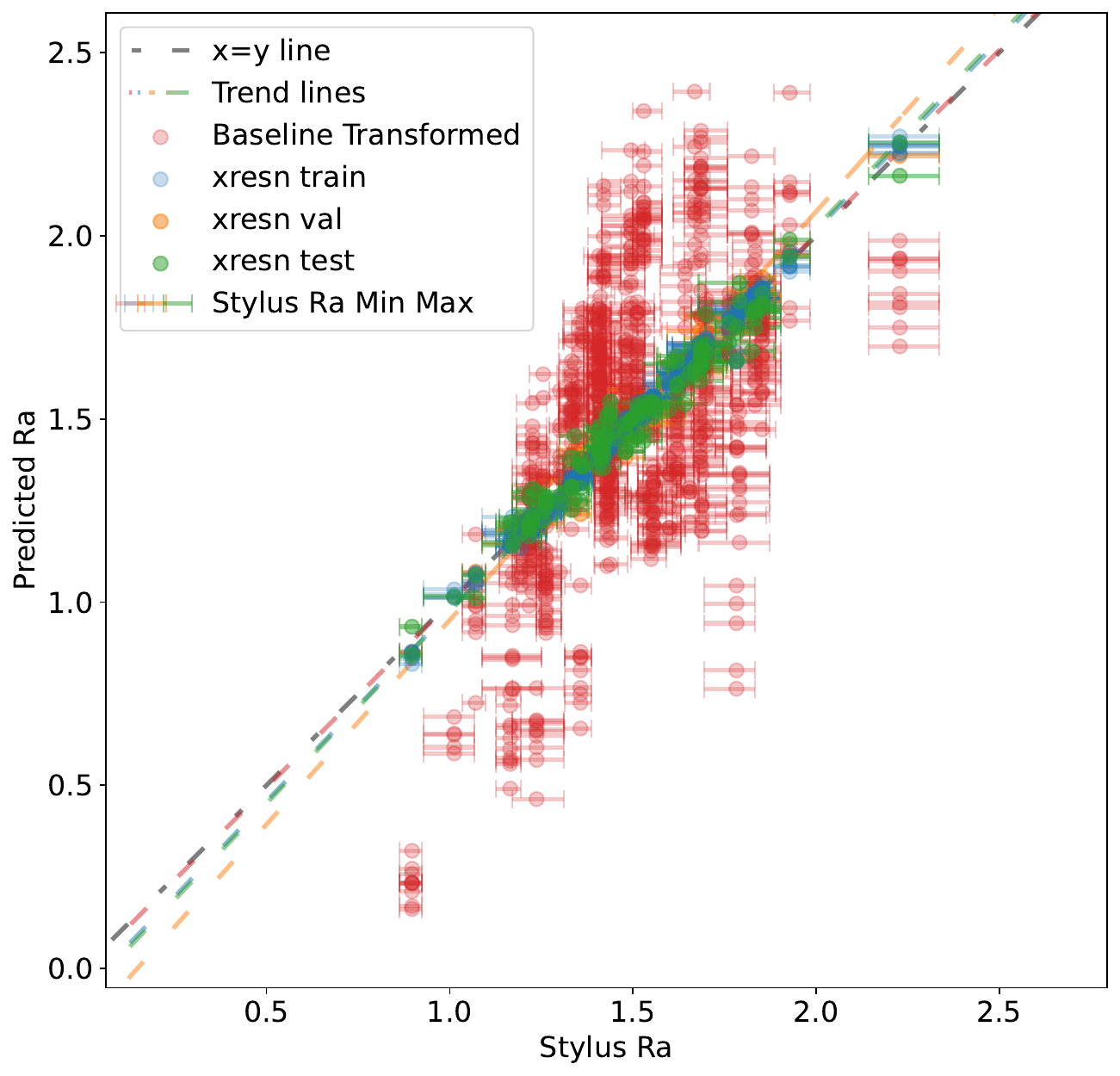}
\caption{The 2D xresnet34 results on the 20\% split dataset vs the baseline tansformed.}

\label{fig:best_20pc_scatter}
\end{figure}

\textbf{2D Deep Learning (Data-Driven Approach)} models showed some variation in performance. It is important to note that the 2D Deep Learning models can be categorized into two groups: convolution-based models (such as xresnet and ConvNeXt) and transformer-based models (such as Swin and ViT). The best-performing model, xresnet34, was better than any of the 1D models, though many of the transformer-based 2D models performed worse than all of them. In the cases where a 2D model corresponded to a 1D architecture, the 2D model performed better. For example, this was the case with the xresnet models. The xresnet34 model achieved the lowest RMSE of 0.0453, with a high correlation coefficient of 0.9806. 
Similar performance was observed for the ConvNeXt Small and ConvNeXt Tiny models, which achieved RMSE values of 0.0554 and 0.0561 respectively.
The xresnet18 model also exhibited relatively strong performance. 
The transformer-based models exhibited relatively lower performance, as indicated by higher RMSE values and lower correlation coefficients compared to the convolution-based models. 

Overall, the best-performing model is the 2D xresnet34. It exhibited the lowest RMSE, highest correlation coefficient, lowest maximum error, and best prediction coverage.
This is closely followed by the best 1D convolution-based TCN model with similar performance but with the benefit in terms of computational efficiency due to less complex 1D convolutional operations. 



The prediction scatter for the 2D xresnet34 on the $20\%$ data is shown in Fig.~\ref{fig:best_20pc_scatter}. It is the best deep learning model using the data-driven approach and significantly improves Ra predictions when compared to the transformed closed-form baseline.
Prediction coverage (Pred. Coverage), is a measure of the proportion of predictions that are within the minimum and maximum of Ra measurements from the stylus for the steel sample. This is visualized in the scatter plots: when a point's error bars cross the x=y line, it is counted as correct for the coverage percentage.

\subsection{K-Fold Experiments}

\begin{figure}[t]
\centering
\includegraphics[width=\linewidth]{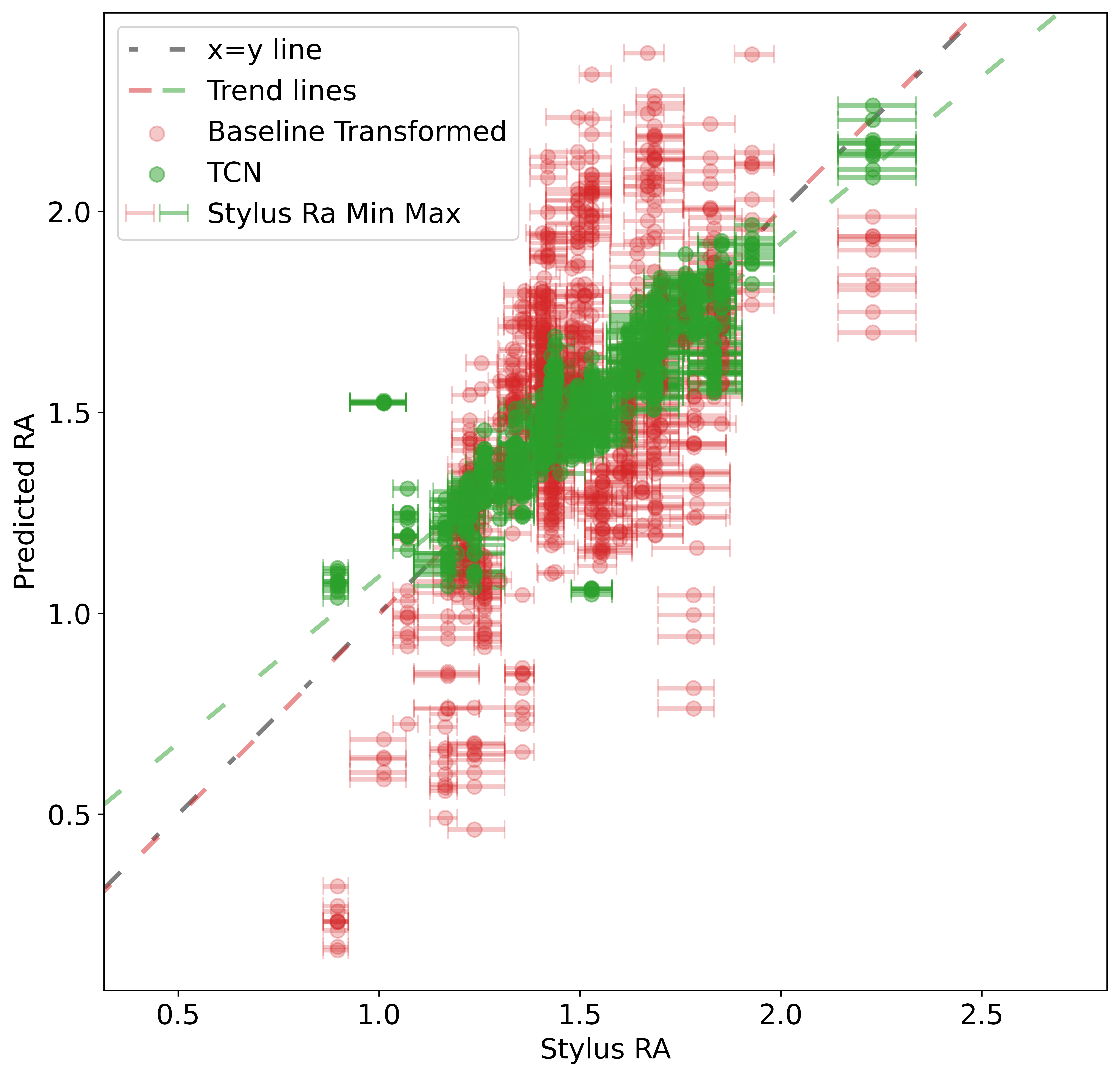}
\caption{The results for the best model on the kfold data (TCN ks=5 nl=13) vs the baseline.}

\label{fig:best_fold_scatter}
\end{figure}

To test the models' generalizability between data samples from different steel samples, k-fold experiments are performed where all samples from each steel sample become the test set for a model trained for each. The results of the k-fold experiments are presented in Table~\ref{tab:fold_model_comparison}. These experiments are significantly more expensive to run, as each model needs to be retrained 49 times, once for each steel sample. Therefore, these experiments have fewer models tested. The baselines are the same for both experiments as they are closed form.

\begin{table*}[ht]
\centering
\caption{Results on Data: Fold Each Steel Sample}
\begin{tabular}{|c|c|c|c|c|}
\hline
\textbf{Model} & \textbf{RMSE} & \textbf{Correlation} & \textbf{Max Error} & \textbf{Pred. Coverage} \\
\hline
\multicolumn{5}{|c|}{\textbf{Baselines}} \\
\hline
\textbf{Baseline Transformed} & 0.3076 & 0.5940 & 1.0193 & 0.0992\\
Baseline & 0.6546 & 0.5940 & 1.2648 & 0.0064\\
\hline
\multicolumn{5}{|c|}{\textbf{Non-Deep Learning (Data-Driven Approach)}} \\
\hline
\textbf{MiniRocket} & 0.1157 & 0.8675 & 0.3944 & 0.3587\\
Rocket & 0.1427 & 0.7861 & 0.4852 & 0.2813\\
\hline
\multicolumn{5}{|c|}{\textbf{1D Deep Learning (Data-Driven Approach)}} \\
\hline
\textbf{TCN ks=5 nl=13} & 0.1028 & 0.8923 & 0.5167 & 0.4542\\
TCN ks=7 nl=8 & 0.1036 & 0.8917 & 0.4902 & 0.4361\\
xresnet1d34 & 0.1076 & 0.8819 & 0.5649 & 0.4284\\
xresnet1d18 & 0.1102 & 0.8784 & 0.6172 & 0.4284\\
TCN ks=9 nl=12 & 0.1109 & 0.8764 & 0.5412 & 0.3832\\
InceptionTime & 0.1115 & 0.8717 & 0.5904 & 0.4503\\
MLSTM-FCN & 0.1174 & 0.8602 & 0.6120 & 0.3729\\
xresnet1d50 & 0.1204 & 0.8490 & 0.6394 & 0.3768\\
MGRU-FCN & 0.1224 & 0.8512 & 0.5105 & 0.3200\\
MRNN-FCN & 0.1267 & 0.8365 & 0.5793 & 0.3561\\
FCN & 0.1359 & 0.8041 & 0.5446 & 0.3006\\
\hline
\multicolumn{5}{|c|}{\textbf{2D Deep Learning (Data-Driven Approach)}} \\
\hline
\textbf{xresnet34} & 0.1095 & 0.8774 & 0.6010 & 0.4181\\
\hline
\end{tabular}
\label{tab:fold_model_comparison}
\end{table*}

The k-fold experiments evaluate the performance of the models on each steel sample individually. In these experiments, the Baseline Transformed and Baseline models serve as the reference baselines for comparison. The Baseline Transformed model achieves an RMSE value of 0.3076 and a correlation coefficient of 0.5940. The Baseline model performs slightly worse, with an RMSE value of 0.6546 and the same correlation coefficient.

\begin{figure}[t]
\centering
\includegraphics[width=\linewidth]{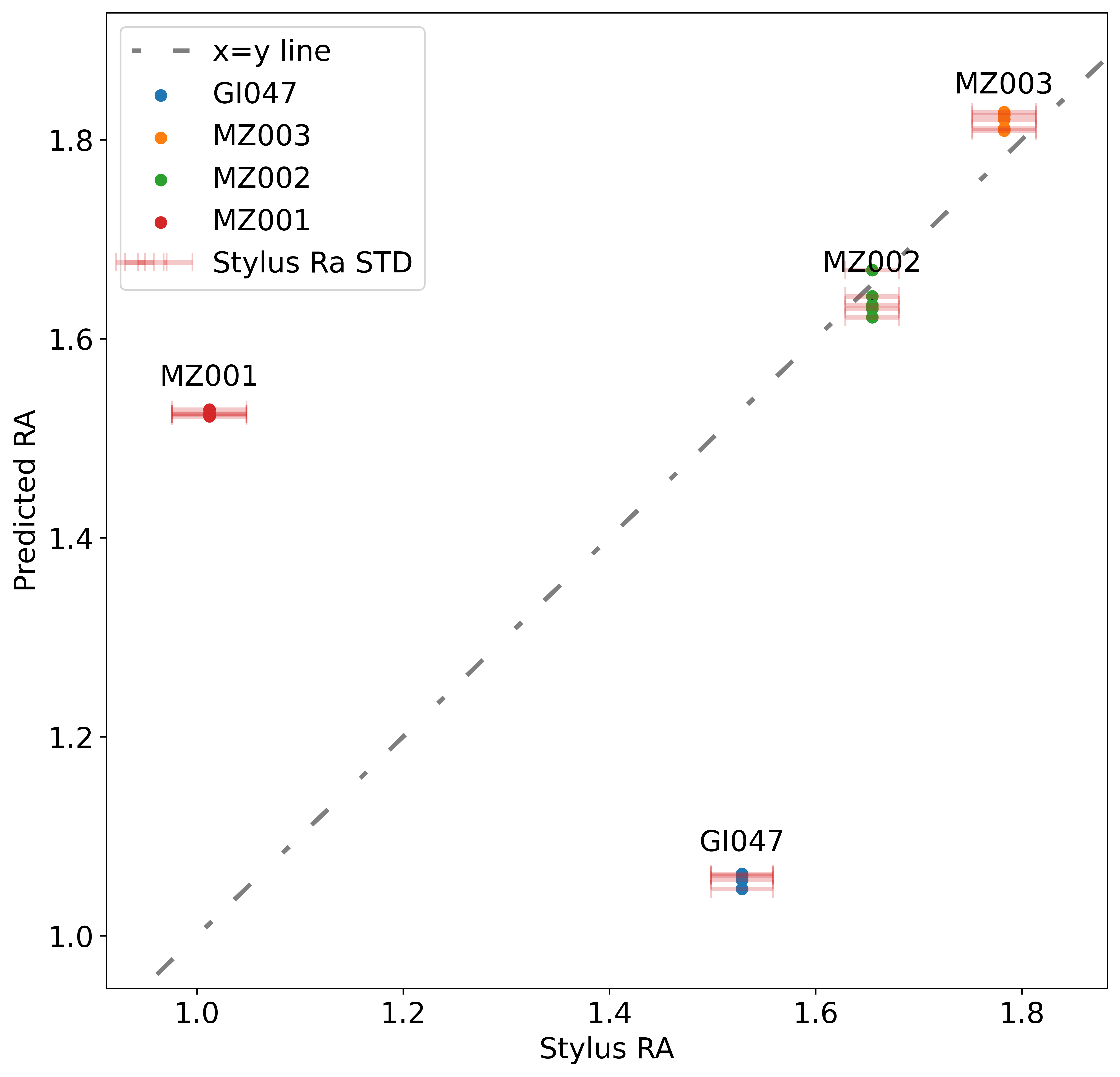}
\caption{The TCN
ks=5 nl=13 model results scatter plot for all of the steel samples that were outliers when we calculated the close form results, namely, all of the ``MZ" steel samples and the ``GI047" steel sample.}

\label{fig:best_fold_scatter_outliers}
\end{figure}

Among the non-deep learning models in the data-driven approach category, the MiniRocket model demonstrates the best performance in the k-fold experiments. It achieves an RMSE value of 0.1157 and a high correlation coefficient of 0.8675. The Rocket model also performs reasonably well, with an RMSE of 0.1427 and a correlation coefficient of 0.7861.

In the 1D deep learning models, the TCN ks=5 nl=13 and TCN ks=7 nl=8 models exhibit strong performance. Both models achieve relatively low RMSE values of 0.1028 and 0.1036, respectively, with high correlation coefficients of 0.8923 and 0.8917.
The 2D xresnet34, achieves an RMSE value of 0.1095 and a correlation coefficient of 0.8774. This is slightly worse than its 1D equivalent, unlike the 20\% split results.

Comparing the performance of the models in the k-fold experiments, it can be observed that the TCN models, lower capacity 1D xresnet models (xresnet1d34, xresnet1d18) models, and 2D xresnet34 outperform the other models. These models perform well across both experiments, with the TCN models constantly performing very well and very significantly beating the baselines. 


As expected, the models perform worse on the k-fold experiments as each prediction is on a completely unseen steel sample, and our dataset has only 49 steel samples, a relatively small number for generalization across samples. 

\begin{figure}[ht]
\centering
\includegraphics[width=\linewidth]{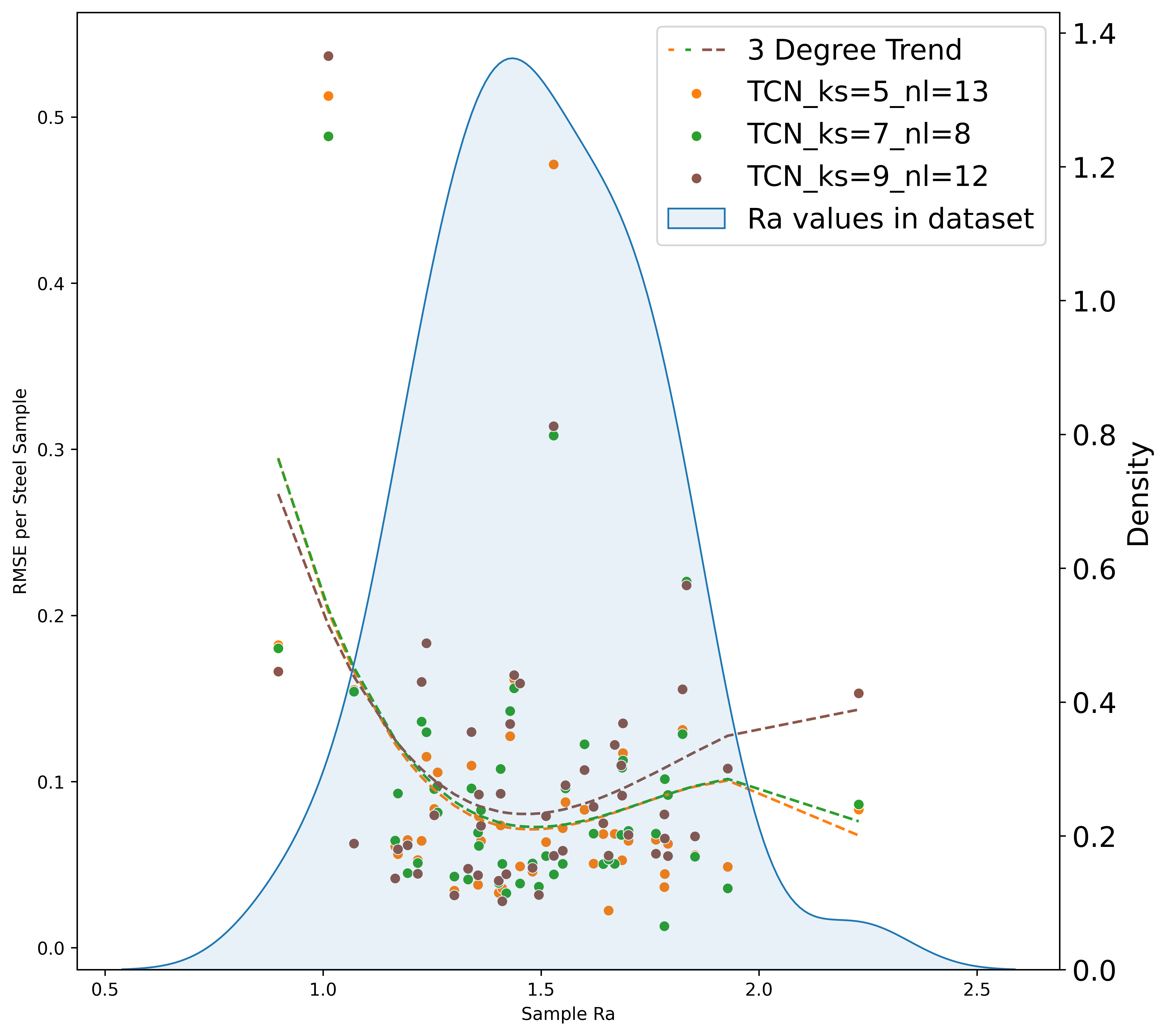}
\caption{The TCN models' results vs the stylus steel sample Ra for each fold. The right axis also shows the density of the Ra values in the dataset.}

\label{fig:kfold_vs_stylus_TCN_with_trend_and_density}
\end{figure}

Fig.~\ref{fig:dist_of_ra} shows that the data has close to a normal distribution of Ra values across our samples. The model has seen more samples with Ra values closer to the mean across all samples. Fig.~\ref{fig:kfold_vs_stylus_TCN_with_trend_and_density} shows the RMSE for each steel sample for the TCN models for the k-fold evaluation. It plots the results vs the stylus measurement for each of the steel samples. 
There appears to be a minor improvement in the results nearer to the distribution mean.

It can be observed that in the k-fold experiments, all the models exhibited poor performance on all of the samples taken from two specific steel samples, namely ``MZ001'' and ``GI047''. Interestingly, these two samples were also identified as outliers in our baseline calculations, as depicted in Fig.~\ref{fig:sorm_vs_our_Ra_calc}. This finding raises the possibility that the models are struggling to accurately predict these results due to inherent issues with the collected data for these samples or problems arising from our data preprocessing procedures. It is important to note that these preprocessing steps are conducted prior to both our calculation of the closed-form solution and the data-driven approach. However, it is worth mentioning that the data-driven approach produced satisfactory results for the other two outliers from the closed-form calculation, namely ``MZ002" and ``MZ003". Fig.~\ref{fig:best_fold_scatter_outliers} shows the TCN
ks=5 nl=13 results scatter plot of only the samples from the closed-form outlier steel samples, ``MZ001", ``MZ002", ``MZ003", and ``GI047". 

Despite these issues, the data-driven approach consistently beats the baseline across all metrics. It is also worth noting that the CNN models are computationally efficient and can be run in real-time for predictions during production.

We hypothesize that the underperformance of transformer approaches can be attributed to the limited size of the available data and the absence of unsupervised pretraining techniques or transfer learning. Typically, transformer networks are trained on significantly larger datasets, such as being pre-trained on extensive corpora and subsequently fine-tuned on the specific experimental dataset.
In our study, we did not employ any form of pretraining for the transformer-based approaches, which likely explains their inferior performance compared to the Cov-based models. Cov-based models possess greater inductive bias, enabling them to learn effectively even with less data. On the other hand, transformer models have the potential to achieve superior results by adopting diverse learning strategies but require a larger amount of data to acquire this capability.

\section{Conclusion}\label{sec:conclusion}



In this paper, we have provided a comprehensive evaluation of machine learning models for the prediction of the Ra roughness parameter from laser light reflection data.

This research introduces a novel methodology that substantially enhances the accuracy of online Ra roughness parameter prediction by using machine learning models—an approach previously unexplored. Conventional stylus measurements cannot be taken on-line, and existing on-line methods have been measured to exhibit inaccuracies compared to the industry-standard stylus measurements. Our approach and the models we tested consistently outperformed the conventional closed-form baseline method, underscoring their effectiveness.  

Moreover, we validated the robustness of our approach through two distinct experimental setups: the 20\% test data experiments and the more challenging k-fold cross-validation. The findings from these experiments highlight not only the potential but also the reliability of our machine learning-based approach, reaffirming its capability to generalize and deliver accurate Ra roughness predictions on unseen samples.

This approach offers fresh insights into how other surface roughness parameters can be predicted and measured. The success of our approach suggests that, in future work, similar models can be employed to predict a spectrum of other steel surface parameters, thereby unlocking a broader range of applications in various domains. By identifying the most effective machine learning models for our specific problem, this work offers a valuable starting point for those encountering similar data challenges.


In summary, this research signifies how accurate machine learning can be for the prediction of surface roughness parameters. Our findings not only emphasize the potential of data-driven approaches but also provide a roadmap for future endeavors in the arena of steel surface parameter prediction, encouraging ongoing exploration and innovation.


\bmhead{Supplementary information}

The code for the experiments is available on github: \url{https://github.com/alexander-milne/ra_roughness_prediction}

\bmhead{Acknowledgments}

The authors would like to thank our industrial partner for providing the data, assistance, and advice.

\bmhead{Funding}
This work was supported by the UKRI (Grant number EP/V519601/1). 

\section*{Declarations}

\bmhead{Conflict of interest} 
The authors have no relevant financial or non-financial interests to disclose.




\end{document}